\documentclass[10pt,conference,letterpaper]{IEEEtran}
\IEEEoverridecommandlockouts

\usepackage{fontspec}
\usepackage{newunicodechar}
\usepackage{pifont}
\newfontfamily{\symfallback}{LibertinusMath-Regular.otf}[Scale=MatchLowercase]
\newunicodechar{≥}{{\symfallback ≥}}
\newunicodechar{≤}{{\symfallback ≤}}
\newunicodechar{≈}{{\symfallback ≈}}
\newunicodechar{≠}{{\symfallback ≠}}
\newunicodechar{≫}{{\symfallback ≫}}
\newunicodechar{≪}{{\symfallback ≪}}
\newunicodechar{∈}{{\symfallback ∈}}
\newunicodechar{∉}{{\symfallback ∉}}
\newunicodechar{⊂}{{\symfallback ⊂}}
\newunicodechar{⊃}{{\symfallback ⊃}}
\newunicodechar{√}{{\symfallback √}}
\newunicodechar{∑}{{\symfallback ∑}}
\newunicodechar{∏}{{\symfallback ∏}}
\newunicodechar{≻}{{\symfallback ≻}}
\newunicodechar{≺}{{\symfallback ≺}}
\newunicodechar{⁻}{{\symfallback ⁻}}
\newunicodechar{⁰}{{\symfallback ⁰}}
\newunicodechar{¹}{{\symfallback ¹}}
\newunicodechar{²}{{\symfallback ²}}
\newunicodechar{³}{{\symfallback ³}}
\newunicodechar{⁴}{{\symfallback ⁴}}
\newunicodechar{⁵}{{\symfallback ⁵}}
\newunicodechar{⁶}{{\symfallback ⁶}}
\newunicodechar{⁷}{{\symfallback ⁷}}
\newunicodechar{⁸}{{\symfallback ⁸}}
\newunicodechar{⁹}{{\symfallback ⁹}}
\newunicodechar{✓}{\ding{51}}
\newunicodechar{✗}{\ding{55}}
\newunicodechar{→}{{\symfallback →}}
\newunicodechar{←}{{\symfallback ←}}
\newunicodechar{α}{{\symfallback α}}
\newunicodechar{β}{{\symfallback β}}
\newunicodechar{γ}{{\symfallback γ}}
\newunicodechar{δ}{{\symfallback δ}}
\newunicodechar{θ}{{\symfallback θ}}
\newunicodechar{κ}{{\symfallback κ}}
\newunicodechar{λ}{{\symfallback λ}}
\newunicodechar{μ}{{\symfallback μ}}
\newunicodechar{π}{{\symfallback π}}
\newunicodechar{ρ}{{\symfallback ρ}}
\newunicodechar{σ}{{\symfallback σ}}
\newunicodechar{τ}{{\symfallback τ}}
\newunicodechar{φ}{{\symfallback φ}}
\newunicodechar{χ}{{\symfallback χ}}
\newunicodechar{ψ}{{\symfallback ψ}}
\newunicodechar{ω}{{\symfallback ω}}
\newunicodechar{Δ}{{\symfallback Δ}}
\newunicodechar{Θ}{{\symfallback Θ}}
\newunicodechar{Σ}{{\symfallback Σ}}
\newunicodechar{Π}{{\symfallback Π}}
\newunicodechar{Λ}{{\symfallback Λ}}
\newunicodechar{Ω}{{\symfallback Ω}}
\newunicodechar{×}{{\symfallback ×}}

\providecommand{\tightlist}{\setlength{\itemsep}{0pt}\setlength{\parskip}{0pt}}

\providecommand{\texorpdfstring}[2]{#1}
\usepackage{amsmath}
\usepackage{amssymb}
\usepackage{booktabs}
\usepackage{calc}
\usepackage{array}
\usepackage{xurl}
\usepackage{url}
\usepackage{graphicx}
\usepackage{etoolbox}
\AtBeginEnvironment{thebibliography}{\footnotesize\sloppy\raggedright}
\renewcommand{\footnoterule}{%
  \kern -3pt
  \hrule width .38\columnwidth height .4pt
  \kern 2.6pt
}

\makeatletter
\def\abstract{\normalfont
    \if@twocolumn
      \@IEEEabskeysecsize\mdseries\textit{\abstractname}---\relax
    \else
      \bgroup\par\addvspace{0.5\baselineskip}\centering\vspace{-1.78ex}\@IEEEabskeysecsize\abstractname\par\addvspace{0.5\baselineskip}\egroup\quotation\@IEEEabskeysecsize
    \fi\@IEEEgobbleleadPARNLSP}
\makeatother



\providecommand{\citep}[2][]{\cite{#2}}
\providecommand{\citet}[2][]{\cite{#2}}

\begin{document}

\title{ValueBlindBench: Agreement-Gated Stress Testing of LLM-Judged
Investment Rationales Before Returns Are
Observable\thanks{ValueBlindBench is independent research by Blossom AI
and is not affiliated with any external company or commercial product.
Earlier public versions of this manuscript may retain a prior name as
part of persistent scholarly records; the current version and
author-controlled materials use the revised name.}}

\author{
\IEEEauthorblockN{Sidi Chang\IEEEauthorrefmark{1},
Peiying Zhu\IEEEauthorrefmark{2}, and
Yuxiao Chen\IEEEauthorrefmark{2}}
\IEEEauthorblockA{\IEEEauthorrefmark{1}Blossom AI Labs, Tokyo, Japan\\
\IEEEauthorrefmark{2}Blossom AI, San Francisco, USA}
}

\maketitle

\begin{abstract}
LLM-based financial agents increasingly produce investment rationales
before the outcomes needed to evaluate them are observable. This creates
a delayed-ground-truth evaluation problem: realized returns remain the
eventual arbiter of investment quality, but they arrive too late and are
too noisy to guide many model-development and governance decisions. LLM
judges offer a tempting shortcut for pre-deployment evaluation of
AI-finance systems, but unvalidated judges may reward verbosity,
confidence, or rubric mimicry rather than financial judgment. This paper
introduces ValueBlindBench, a preregistered agreement-gated stress-test
protocol for deciding when LLM-judged investment-rationale claims are
publishable, qualified, or invalid.

In a controlled market-state capital-allocation prototype with 1,000
honest decision cycles and 100 preregistered adversarial controls (1,100
trajectories, 5,500 judge calls), ValueBlindBench clears the aggregate
agreement gate at \(\bar{\kappa}_w = 0.7168\) but prevents several
overclaims. Lower-rank systems collapse into a tie-class, one rubric
dimension fails the per-dimension gate (\texttt{constraint\_awareness},
\(\bar{\kappa}_w = 0.2022\)), single-judge rankings are
family-dependent, and terse-correct rationales receive a
\(\Delta = -2.81\) rubric-point penalty relative to honest rationales. A
targeted anchor-specificity probe further shows that financial
constructs such as constraint awareness are operationally load-bearing.

The scientific object is therefore not a leaderboard and not a claim to
measure true investment skill. ValueBlindBench is a pre-calibration
metrology layer for AI-finance evaluation: it governs whether a proposed
LLM-judge-based investment-rationale claim is stable enough, agreed
enough, and uncontaminated enough to be reported at all.
\end{abstract}

\section{Introduction}\label{introduction}

\subsection{The pre-realization problem in
finance}\label{the-pre-realization-problem-in-finance}

Recent advances in LLM-based financial agents make it increasingly easy
to generate plausible investment rationales at scale. Evaluation has not
kept pace. In capital-allocation settings, the rationale is available
immediately, but the outcome that would validate or falsify the decision
may arrive months or years later, and even then may be confounded by
market noise. A value-investing thesis may look wrong for two quarters
and be correct over five years; a short-horizon trade can profit by
accident. We call this the pre-realization evaluation problem.

This paper studies that problem for AI trading agents,
capital-allocation systems, and financial analytics workflows where
model-development and governance decisions must be made before realized
returns become informative. The core question is not whether an AI
system is a good investor. It is a prior measurement question: when is
an LLM-judge-based evaluation claim safe to report? If pre-realization
claims are made without measurement discipline, the field risks
selecting systems for surface plausibility rather than for robust
decision support.

The work sits within a broader financial AI metrology agenda: before
financial AI evaluation outputs are used for model selection,
deployment, or governance, the measurement instrument itself must be
audited. ValueBlindBench addresses the delayed-ground-truth regime. In
supervised financial NLP, gold labels may exist but rubric and metric
choices still require audit. In pre-realization investment evaluation,
the problem is harder: outcome labels arrive late, are noisy, and cannot
support immediate governance decisions. ValueBlindBench therefore
governs claims through agreement, stability, and adversarial gates.

We use long-horizon investment as the proving ground because it offers
four properties the metrology problem requires: decision-time rationales
with controlled inputs, preregisterable regimes, numerically explicit
constraints against which a rubric anchor like \emph{constraint
awareness} can be operationalized, and a cost envelope that fits within
research budgets at the panel sizes the protocol requires. We make the
substrate scope honest at first appearance: this is a market-state
capital-allocation prototype, not a fundamental-equity memo task. What
transfers between substrates is the measurement problem; what does not
transfer without further work is the asset class, horizon, or input
modality.

We do not treat human expert labels as ground truth. In pre-realization
investment settings, experts face the same delayed-outcome problem as
models: they may disagree, overweight presentation quality, and be
unable to distinguish good decision-time judgment from lucky or unlucky
future realization. Human panels are therefore not a privileged validity
source; they are another measurement instrument whose agreement,
stability, and adversarial robustness must also be tested.

\subsection{Why ``just use LLM judges'' is not
enough}\label{why-just-use-llm-judges-is-not-enough}

LLM judges are an attractive candidate for the pre-realization interval:
cheap relative to expert review, fast relative to realized-outcome
horizons, consistent in formatting. But unvalidated LLM judges are
themselves a measurement instrument with unknown calibration. The
literature documents that they reward longer rationales, more confident
phrasing, and rubric-mimicking text that resembles the kind of memo on
which they were post-trained, regardless of whether the underlying
judgment is sound. Substituting an unvalidated LLM judge for an
unavailable outcome therefore replaces a known noise source with an
unknown bias.

The right response is not to claim that LLM judges are valid. It is to
build a measurement protocol that \emph{decides when their verdicts are
publishable, qualified, or invalid}. This paper addresses a prior
measurement question. Before asking whether an LLM judge agrees with
realized returns, expert committees, or institutional preferences, we
ask whether the judge instrument itself can detect its own instability,
judge-family dependence, anchor ambiguity, and adversarial construct
contamination. The protocol must be able to refuse to issue a
leaderboard when its own preconditions fail.

\subsection{ValueBlindBench}\label{valueblindbench}

We introduce ValueBlindBench, a preregistered agreement-gated
stress-test protocol for LLM-judged AI investment rationales.
ValueBlindBench addresses the prior measurement question rather than the
investment-performance question. It does not validate investment skill
and does not replace realized returns. Instead, it audits the judge
instrument itself: whether judges agree, whether rankings are stable
across judge families, whether dimensions are reliable, and whether
adversarial rationales expose construct contamination. Its output is not
a score or leaderboard, but a verdict on whether a proposed
LLM-judge-based investment-rationale claim is publishable, qualified, or
invalid before returns are observable.

Formally, that verdict is represented as a tuple specifying, for each
candidate claim, the level at which the claim is permitted to be made.

\subsection{Contributions}\label{contributions}

We make five contributions.

Contribution 1 --- problem formulation. We formalize the pre-realization
evaluation problem as a named measurement-discipline question for AI
investment rationales: how should an evaluation claim be governed when
the returns that would adjudicate investment quality are unavailable at
decision time? The term is our shorthand for the interval in which
process and construct validity must be assessed before outcome validity
is observable.

Contribution 2 --- protocol. We introduce ValueBlindBench, a
preregistered, agreement-gated claim-permission protocol whose
deliverable is a verdict tuple specifying claim scope, agreement status,
stability status, adversarial status, and permitted publication level,
rather than a scalar score.

Contribution 3 --- diagnostics. We show how aggregate \(\bar{\kappa}_w\)
gates, per-dimension gates, leave-one-judge-out ranking-stability
checks, single-judge baselines, out-of-family probes, and adversarial
cells isolate the specific claims that ordinary leaderboard reporting
would otherwise overstate.

Contribution 4 --- empirical stress test. We apply the protocol to a
controlled market-state capital-allocation prototype with four frontier
rationale-generating agents across five market regimes (1,000 honest
decision cycles plus 100 preregistered adversarial controls; 1,100
trajectories scored with 5,500 judge calls). The protocol both
authorizes a qualified rank-1 claim and withholds lower-rank and
dimension-level overclaims.

Contribution 5 --- measurement-failure discovery. We demonstrate that
terse-correct rationales are severely penalized under the v1.0 rubric
(\(\Delta = -2.81\) rubric points, with a non-linear threshold near 60
tokens), showing that naive LLM-judge finance evaluation can confuse
substantive adequacy with rhetorical coverage. The same stress test
surfaces single-judge-family arbitrariness, abstract-anchor ambiguity,
and lower-rank tie-classes as first-order threats to LLM-judged
investment-rationale claims.

\emph{The unit of contribution is not a model score but a permissioned
claim: ValueBlindBench determines whether a proposed
investment-rationale claim is publishable, qualified, or invalid under
preregistered measurement rules.}

\section{Related work}\label{related-work}

Financial model evaluation under short-window noise. Realized P\&L on
sub-multi-year windows has signal-to-noise ratios near unity
\citep{lo2002statistics}; backtest overfitting and multiple-testing
error contaminate even longer horizons
\citep{bailey2014pseudomathematics, lopezdeprado2018}. These critiques
are the reason a non-P\&L evaluation layer is required in the
pre-realization interval. They do not argue that P\&L is wrong; they
argue that P\&L is \emph{insufficient} on the timescales at which
model-development decisions are made.

LLM-as-judge and its documented failure modes. Multi-judge ensembles
have grown since AlpacaEval and MT-Bench
\citep{zheng2023judging, liu2023geval}, while broad evaluation suites
such as HELM establish the closest multi-metric-evaluation lineage
\citep{liang2023helm}. Subsequent work documents verbosity preference,
position bias, self-preference, and stylistic-mimicry effects
\citep{hosking2023}, among others. ValueBlindBench's primary-protocol
components are designed as structural responses to this threat catalog:
the adversarial control cells expose verbosity and confidence biases,
the LOFO check exposes single-judge-family preference, and the
bifurcated \(\bar{\kappa}_w\) gate refuses publication when the panel
fails to converge. ValueBlindBench is therefore not another LLM-as-judge
benchmark. It is a claim-permission protocol for financial decision
rationales under delayed outcomes.

Measurement validity, metrology, and preregistration. ValueBlindBench
treats judge evaluation as a construct-validity problem in the sense of
Cronbach--Meehl and Messick: the protocol does not ask only whether a
score is reproducible, but what claim the score is evidence for
\citep{cronbach1955construct, messick1989validity}. The bifurcated
\(\bar{\kappa}_w\) gate operationalizes a Krippendorff-style discipline
\citep{krippendorff2018} that an inter-rater agreement statistic should
be reported with a decision rule, not interpreted post hoc against a
free-floating threshold; we explicitly reject the
unweighted-Fleiss-\(\kappa\)-with-Landis--Koch labels practice
\citep{landis1977} as the gating statistic on a 5-level ordinal rubric.
Preregistration is increasingly advocated in behavioral science
\citep{nosek2018preregistration}, while adjacent computational fields
continue to document weak reproducibility and sharing practices
\citep{wieling2018}; our protocol locks the analysis plan before any
data are collected.

The relevant comparison for this paper is not ``LLM judge versus human
ground truth.'' In pre-realization finance, expert panels are themselves
measurement instruments: they may disagree, display institutional style
preferences, and remain unable to know true long-horizon outcome quality
at decision time. The comparison ValueBlindBench makes available is
between measurement instruments with gates and measurement instruments
without gates.

\section{The ValueBlindBench
protocol}\label{the-valueblindbench-protocol}

\subsection{Object of evaluation}\label{object-of-evaluation}

ValueBlindBench stress-tests an LLM-judge measurement instrument over
\emph{observable investment rationales under a fixed rubric and a fixed
judge panel}. It does not directly observe latent model reasoning;
rationales are post-hoc generations that may or may not reflect the
agent's internal decision process. It does not replace realized
outcomes; the eventual arbiter of investment quality remains realized
P\&L on sufficiently long horizons. What it does claim is earlier in the
measurement pipeline: \emph{given} that some pre-realization evaluation
must occur to support model-development and governance decisions, a
preregistered agreement-gated stress test can determine whether an
LLM-judge-based claim is agreed enough, stable enough, and
uncontaminated enough to be reported.

\subsection{Substrate: a controlled market-state prototype, not full
value
investing}\label{substrate-a-controlled-market-state-prototype-not-full-value-investing}

The empirical substrate of this paper is a controlled market-state
capital-allocation prototype, not a full company-fundamental
value-investing memo task. Each cycle presents the rationale-generating
agent with 60 sanitized one-minute price bars, a portfolio state, and a
constraint set; the agent issues a small number of trade decisions and a
free-text rationale.

We use this substrate because it offers four properties the metrology
problem requires: decision-time rationales with controlled inputs,
preregisterable market regimes, numerically explicit portfolio
constraints against which an anchor like \emph{constraint awareness} can
be operationalized, and a cost envelope that fits within research
budgets at the panel sizes the protocol requires. It does \emph{not}
offer fundamental-memo realism: no SEC filings, no qualitative
disclosures, no multi-quarter horizons.

What transfers between this prototype and fundamental-equity-memo
settings is the \emph{measurement problem}: evaluating rationales before
outcomes become informative. What does \emph{not} transfer without
further work is the asset class, horizon, or input modality. A future
preregistered extension should test the protocol on a fundamental-memo
substrate; the present paper claims the protocol, not the extension.

\subsection{Composer panel}\label{composer-panel}

Four rationale-generating agents from distinct provider families, chosen
to bound any single training-corpus bias on the headline ranking.
Provider-side identifiers and release dates are recorded in every result
row's \texttt{composer\_system\_fingerprint} field for reproducibility.

{\def\LTcaptype{none} 
\begin{table*}[!ht]
\centering
\scriptsize
\setlength{\tabcolsep}{3pt}
\begin{tabular}{@{}>{\raggedright\arraybackslash}p{0.23\textwidth}>{\raggedright\arraybackslash}p{0.16\textwidth}>{\raggedright\arraybackslash}p{0.12\textwidth}>{\raggedright\arraybackslash}p{0.39\textwidth}@{}}
\toprule\noalign{}
Agent ID & Provider & Released & API path \\
\midrule\noalign{}
\texttt{gpt-5.5} & OpenAI & 2026 Q1 & OpenAI Batch API \\
\texttt{claude-sonnet-4-6} & Anthropic & 2026 Q1 & Anthropic Message
Batches \\
\texttt{gemini-3.1-pro-preview} & Google & 2026 Q1 & Google AI Studio
Batch \\
\texttt{qwen3-235b-a22b} & Alibaba via Vertex MaaS & 2025-07 & Vertex
MaaS direct async \\
\end{tabular}
\end{table*}
}

\subsection{Judge ensemble}\label{judge-ensemble}

Three judges with five calls per trajectory: Claude Sonnet 4.6
(\(\times 3\) no-seed trials, averaged at the (decision, judge) level
before downstream analysis); GPT-5.5 (\(\times 1\), seed=42, reasoning
effort high); Gemini 3.1 Pro Preview (\(\times 1\), temperature 0).
Within-judge Repetition Stability on Claude is required to clear
\(0.90\) before trial averaging is treated as defensible. Repetition
Stability is computed on aggregate rubric scores as
\(RS = 1 - \bar{\sigma}^2_{\text{within-trial}} / \sigma^2_{\text{all trials}}\),
where \(\bar{\sigma}^2_{\text{within-trial}}\) is the mean variance of
the three Claude trials within each scored trajectory and
\(\sigma^2_{\text{all trials}}\) is Claude's total variance across all
trial-level aggregate scores; the \(0.90\) gate was preregistered. RS is
a preregistered variance-ratio repetition-stability diagnostic, not an
ICC(1): it is used only to decide whether averaging the three Claude
trials is defensible, not to estimate a population random-effects
reliability coefficient. In the present run, Claude Sonnet 4.6 clears
this gate with Repetition Stability \(=0.9874\); GPT-5.5 and Gemini
contribute single trials, so repetition stability is undefined for those
judges.

\subsection{Adversarial control cells}\label{adversarial-control-cells}

Two preregistered adversarial cells of 50 trajectories each test
specific construct contaminations.

\begin{itemize}
\tightlist
\item
  Cell A (verbose-confident-but-wrong). Preregistered prediction: scores
  in the bottom quartile of the honest distribution at \(p < 0.01\).
\item
  Cell B (terse-but-correct, \(\leq 60\) tokens, \(\geq 3\) features
  cited, no hedging). Preregistered \emph{trichotomy}: H1 (substance)
  within \(\pm 0.3\) of honest; H1' (verbosity bias) more than \(0.5\)
  below honest with multiple-comparison correction; H1 double-prime
  (inconclusive) between \(0.3\) and \(0.5\) below.
\end{itemize}

A LOFO halo check, locked as a primary-protocol amendment, drops the
in-family judge for each cell and recomputes the score; a halo of
\(|\Delta_{\text{full}} - \Delta_{\text{LOFO}}| \geq 0.3\) rubric points
triggers the LOFO score becoming primary for that cell. The same locked
amendment also commits, in advance, to running an out-of-family
fourth-judge probe if the LOFO ranking-stability check returns
\(\rho < 0.9\) on any single drop.

\subsection{\texorpdfstring{Aggregate \(\bar{\kappa}_w\)
definition}{Aggregate \textbackslash bar\{\textbackslash kappa\}\_w definition}}\label{aggregate-barkappa_w-definition}

Within-judge trial averaging is applied first: for the multi-trial judge
(Anthropic, \(\times 3\) trials per (cycle, judge)), the three scores
are averaged to one rater-score per pair. For \(\kappa\) calculations
only, this trial-averaged score is rounded to the nearest rubric
category and clipped to the 1--5 ordinal scale; score means and
effect-size analyses retain floating-point averages. Pairwise
quadratic-weighted Cohen's \(\kappa\) is then computed for each of the
\(C(3,2) = 3\) judge-family pairs on the resulting
one-rater-score-per-judge data. The aggregate \(\bar{\kappa}_w\)
reported throughout this paper is the arithmetic mean of the three
pairwise \(\kappa_w\) values:

\[\bar{\kappa}_w \;=\; \tfrac{1}{3}\!\left[\kappa_w^{(\text{cl},\text{gpt})} \!+\! \kappa_w^{(\text{cl},\text{gem})} \!+\! \kappa_w^{(\text{gpt},\text{gem})}\right].\]

This \(\bar{\kappa}_w\) is a protocol-level summary convention, not a
new three-rater chance-corrected coefficient. We report unweighted
Fleiss \(\kappa\) separately as a transparency statistic, but it is not
the publication gate.

Per-dimension \(\bar{\kappa}_w\) is computed by the same procedure on
each rubric dimension's rounded ordinal scores separately, before the
aggregate-score reduction. We report \(\bar{\kappa}_w\) rather than
unweighted Fleiss \(\kappa\) as the gating statistic because the rubric
is ordinal: a 4-vs-5 disagreement is substantively closer to agreement
than a 1-vs-5 disagreement, and quadratic weighting is the standard
ordinal-aware adjustment.

\subsection{Bifurcated publication
gate}\label{bifurcated-publication-gate}

{\def\LTcaptype{none} 
\begin{table*}[!ht]
\centering
\small
\setlength{\tabcolsep}{3pt}
\begin{tabular}{@{}>{\raggedright\arraybackslash}p{0.20\textwidth}>{\raggedright\arraybackslash}p{0.24\textwidth}>{\raggedright\arraybackslash}p{0.50\textwidth}@{}}
\toprule\noalign{}
Gate outcome & Condition (aggregate \(\bar{\kappa}_w\)) & Permitted claim \\
\midrule\noalign{}
Publish & \(\bar{\kappa}_w \geq 0.4\) & Headline ranking or dimension
claim allowed \\
Methodology finding & \(0.2 \leq \bar{\kappa}_w < 0.4\) &
Judge-convergence or anchor issue reported; no headline claim \\
Halt & \(\bar{\kappa}_w < 0.2\) & No ranking; post-mortem only \\
\end{tabular}
\end{table*}
}

These thresholds are operational decision thresholds, not universal
labels of agreement quality. They were fixed before scoring to prevent
post-hoc threshold tuning and to separate three publication regimes:
sufficient convergence for a scoped claim, partial convergence suitable
only as a methodology finding, and non-convergence requiring halt.

\subsection{The ValueBlindBench
verdict}\label{the-valueblindbench-verdict}

Definition (ValueBlindBench Verdict). For each candidate
investment-rationale claim \(c\), the protocol returns a verdict tuple

\[
\begin{aligned}
V(c)=\langle&
\text{claim\_scope},\\
&\text{agreement\_status},\ \text{stability\_status},\\
&\text{adversarial\_status},\\
&\text{permitted\_publication\_level}\rangle .
\end{aligned}
\]

with the following fields:

{\def\LTcaptype{none} 
\begin{table*}[!ht]
\centering
\small
\setlength{\tabcolsep}{3pt}
\begin{tabular}{@{}>{\raggedright\arraybackslash}p{0.32\textwidth}>{\raggedright\arraybackslash}p{0.62\textwidth}@{}}
\toprule\noalign{}
Field & Domain \\
\midrule\noalign{}
\texttt{claim\_scope} & aggregate ranking; per-dimension ranking;
pairwise contrast; adversarial-cell verdict \\
\texttt{agreement\_status} & publish; methodology; halt \\
\texttt{stability\_status} & stable at claim scope (the stated claim
survives all LOFO drops); tie-class (LOFO failure resolved by
out-of-family probe); judge-dependent (LOFO failure unresolved) \\
\texttt{adversarial\_status} & passed; construct-sensitive;
contaminated; not tested \\
\texttt{permitted\_publication\_level} & headline (all prior fields
green); qualified (methodology or construct-sensitive field); no-claim
(halt, judge-dependent, or same-claim contamination) \\
\end{tabular}
\end{table*}
}

The verdict tuple is the protocol's deliverable. A score-only output is,
in this framework, malformed: it omits the diagnostic context required
to decide whether the score may be acted upon. Each diagnostic
corresponds to a claim that ordinary leaderboard reporting would
otherwise overstate: aggregate agreement governs headline claims,
per-dimension agreement governs construct-level claims, LOFO checks test
judge-family dependence, single-judge baselines expose unstable
shortcuts, and adversarial cells test whether the rubric rewards the
intended financial construct rather than style. Aggregate ranking
claims, per-dimension claims, lower-rank pairwise claims, and
adversarial-cell claims can therefore receive different publication
permissions in the same experimental run.

\subsection{Primary statistical test}\label{primary-statistical-test}

Mean-difference per composer pair across all five regimes,
regime-cluster bootstrap (1,000 resamples, 95\% percentile CI), Holm
step-down over \(C(4,2) \times 6 = 36\) contrasts at \(\alpha = 0.05\).
The post-cutoff regime remains in the paired ranking analysis; it is
also tested separately as the contamination-boundedness probe. The Holm
family consists of the 36 pooled-across-five-regime contrasts; the
post-cutoff probe is a separate boundedness diagnostic and does not add
regime-specific contrasts to the multiplicity family. The pooled minimum
detectable effect of \(0.29\) rubric points is an ex ante planning
calculation using a preregistered paired-difference standard deviation
assumption of \(\sigma_d=1\) at 80\% power; the reported verdicts rely
on the observed bootstrap CIs rather than this planning assumption.

Judge calls are not treated as independent decision observations; the
unit of inference is the decision cycle, equivalently the trajectory,
with repeated calls used for judge aggregation and agreement estimation.

\section{Experimental setup}\label{experimental-setup}

The experiment contains 1,000 honest decision cycles plus 100
preregistered adversarial controls (50 verbose-confident; 50
terse-correct), yielding 1,100 scored trajectories and 5,500 judge calls
(Anthropic \(\times 3\) trials, GPT-5.5 \(\times 1\), Gemini
\(\times 1\)). All five regimes contribute 50 cycles per agent at a
60-bar (1-minute) decision interval. Solana is excluded from the two
earliest regimes for availability reasons. Inputs are sanitized to
remove date and event identity (sanitization procedure in supplementary
material; date-probe acceptance test PASS at protocol-lock).

\subsection{Claim map}\label{claim-map}

{\def\LTcaptype{none} 
\begin{table*}[!ht]
\centering
\scriptsize
\setlength{\tabcolsep}{3pt}
\begin{tabular}{@{}>{\raggedright\arraybackslash}p{0.14\textwidth}>{\raggedright\arraybackslash}p{0.20\textwidth}>{\raggedright\arraybackslash}p{0.36\textwidth}>{\raggedright\arraybackslash}p{0.24\textwidth}@{}}
\toprule\noalign{}
Experiment & What it tests & Result & Permitted conclusion \\
\midrule\noalign{}
Aggregate \(\bar{\kappa}_w\) gate & Do judges converge overall? &
\(0.7168\), 95\% CI \([0.7006, 0.7330]\) & Aggregate analysis
publishable \\
Per-dimension gate & Are all dimensions equally reliable? & 5 pass;
\texttt{constraint\_awareness} = 0.2022, CI \([0.1792, 0.2256]\) & One
dimension downgraded; no headline claim on it \\
Bootstrap rank distribution & Are model ranks stable? & rank-1 stable;
lower-rank bootstrap mass dispersed across qwen3, Gemini, and GPT-5.5 &
Only rank-1 claim publishable \\
LOFO ranking stability & Does rank structure survive judge-family drops?
& drop-Gemini \(\rho=1.0\); drop-GPT-5.5 \(\rho=1.0\); drop-Claude
\(\rho=0.2\) & Rank-1 stable; lower ranks require tie-class \\
Single-judge baseline & Is one judge enough? & rankings differ across
judges, \(\rho = 0.2\)--\(0.8\) & Single-judge protocol insufficient for
lower ranks \\
Cell A & Reject verbose-confident-but-wrong? & panel mean 1.44 vs honest
4.35, \(p < 10^{-3}\) & Panel not simply rewarding verbosity \\
Cell B & Accept correct-but-terse? & mean 1.54 vs honest 4.35,
\(\Delta = -2.81\) & v1.0 is construct-sensitive for terse rationales;
headline ranking qualified \\
v1.1 anchor probe & Is the failing dimension anchor-sensitive? &
\(\Delta = -0.42\) shift; ranking flip & Anchor specificity is
operationally load-bearing; full v1.1 reliability remains future work \\
Out-of-family 4th-judge probe & Scope: lower ranks. Is the LOFO failure
a Claude-family halo? & \(\Delta = +0.04\), CI \([-0.01, +0.09]\) &
Lower ranks confirmed as tie-class, not strict order \\
Repetition stability & Is Anthropic trial averaging defensible? & RS =
0.9874 & Trial averaging clears the 0.90 gate \\
\end{tabular}
\end{table*}
}

The post-cutoff contamination probe (\(\Delta = -0.009\), \(p = 0.86\)
aggregate; non-significant on BTC and SOL splits independently) is
folded into Result 1.

\section{Results}\label{results}

\subsection{Result 1 --- Aggregate agreement is not dimension
validity}\label{result-1-aggregate-agreement-is-not-dimension-validity}

The aggregate \(\bar{\kappa}_w = 0.7168\) clears the publish-headline
gate, with a regime-cluster bootstrap 95\% CI of \([0.7006, 0.7330]\).
The lower bound remains far above the preregistered \(0.4\) publication
threshold. The per-dimension gate, however, classifies one of six
dimensions as a methodology finding rather than a publishable claim:

{\def\LTcaptype{none} 
\begin{table*}[!ht]
\centering
\scriptsize
\setlength{\tabcolsep}{3pt}
\begin{tabular}{@{}>{\raggedright\arraybackslash}p{0.25\textwidth}>{\raggedright\arraybackslash}p{0.13\textwidth}>{\raggedright\arraybackslash}p{0.20\textwidth}>{\raggedright\arraybackslash}p{0.34\textwidth}@{}}
\toprule\noalign{}
Dimension & \(\bar{\kappa}_w\) & 95\% CI & Verdict \\
\midrule\noalign{}
\texttt{action\_coherence} & 0.9354 & \([0.9216, 0.9510]\) & publish \\
\texttt{risk\_alignment} & 0.8834 & \([0.8725, 0.8949]\) & publish \\
\texttt{uncertainty\_handling} & 0.7905 & \([0.7621, 0.8203]\) &
publish \\
\texttt{position\_sizing} & 0.7626 & \([0.7470, 0.7770]\) & publish \\
\texttt{information\_use} & 0.6037 & \([0.5979, 0.6082]\) & publish \\
\texttt{constraint\_awareness} & 0.2022 & \([0.1792, 0.2256]\) &
methodology by the preregistered point-estimate rule \\
\end{tabular}
\end{table*}
}

The aggregate-only verdict would have authorized every per-dimension
claim. The protocol's per-dimension gate prevents this overclaim by
construction. The three Holm-significant pairwise contrasts on
\texttt{constraint\_awareness} are downgraded to \emph{tentative; the
dimension's anchors are the methodology bottleneck}.

The \texttt{constraint\_awareness} interval straddles the \(0.2\) halt
boundary: under the locked point-estimate gate it lands in the
methodology band, while a conservative lower-bound rule would halt that
dimension. The latter rule was not preregistered and is reported only as
a sensitivity disclosure. The Holm family is defined before any
dimension is downgraded: all \(C(4,2) \times 6 = 36\) pairwise
composer-by-dimension contrasts remain in the multiplicity adjustment.
Of the 20 Holm-significant contrasts, three are on
\texttt{constraint\_awareness}; downgrading that dimension changes their
permitted publication level, but does not remove them from the
error-rate family after seeing the data.

We also recomputed the verdict table under stricter operational
publish/methodology/halt cutoffs of \(0.5/0.25\) instead of \(0.4/0.2\).
The aggregate gate remains publishable under this stricter rule, while
\texttt{constraint\_awareness} does not become a headline-publishable
dimension and instead falls to halt/non-claim because even its CI upper
bound is below \(0.25\). Thus the headline rank-1 claim is insensitive
to this stricter threshold, while the already-downgraded dimension
becomes more conservative.

The post-training-cutoff regime returns \(\Delta = -0.009\) rubric
points at \(p = 0.86\) (aggregate), with \(p = 0.55\) on BTC-only and
\(p = 0.35\) on SOL-only splits. The split sample sizes are
\(n_{\text{recent}}=200\), \(n_{\text{historical}}=800\) for BTC and
\(n_{\text{recent}}=200\), \(n_{\text{historical}}=400\) for SOL. The
contamination-bounded interpretation preregistered at protocol-lock is
supported.

\subsection{Result 2 --- Rank-1 is stable, but lower ranks are not
rankable}\label{result-2-rank-1-is-stable-but-lower-ranks-are-not-rankable}

Claude Sonnet 4.6 is the only stable rank-1 claim in this prototype: it
ranks first in 1,000 of 1,000 regime-cluster bootstrap resamples and in
all three single-judge analyses. The out-of-family fourth-judge probe is
scoped to the lower-rank qwen3-vs-gemini tie; it does not certify the
rank-1 headline. The protocol does not authorize a strict ordering of
the remaining systems. The Claude rank-1 result is reported as a worked
example of ValueBlindBench claim permission, not as the paper's primary
scientific object.

Positions 2--4 are not strictly ordered. The full LOFO triplet is:
drop-Gemini \(\rho = 1.0\), drop-GPT-5.5 \(\rho = 1.0\), and drop-Claude
\(\rho = 0.2\). The drop-Claude failure fired the preregistered
out-of-family fourth-judge probe (DeepSeek-V4-Pro). The fourth judge's
verdict on the qwen3 vs gemini gap was \(\Delta = +0.04\) rubric points
(95\% CI \([-0.01, +0.09]\)): not significantly different from zero. In
the main panel, the lower-rank aggregate mean separations are also below
the pooled MDE: GPT-5.5 vs Gemini differs by 0.009, GPT-5.5 vs qwen3 by
0.027, and Gemini vs qwen3 by 0.018 rubric points. Crucially for the
self-preference check, Claude remains rank-1 when Claude-as-judge is
dropped; the drop-Claude \(\rho=0.2\) failure is driven by positions
2--4. Lower ranks form a tie-class.

A financial evaluation protocol should not convert noise-level
differences into artificial precision. The protocol's \emph{non}-result
on positions 2--4 is itself a result.

\subsection{Result 2a --- Worked verdict tuples for this
run}\label{result-2a-worked-verdict-tuples-for-this-run}

Applying the verdict tuple to the present run makes the protocol's claim
permissions explicit:

Worked verdict 1 --- Claude Sonnet 4.6 rank-1. The candidate claim is
scoped to the aggregate ranking. The agreement field is
\texttt{publish}: \(\bar{\kappa}_w = 0.7168\), with 95\% CI
\([0.7006, 0.7330]\). The stability field is
\texttt{stable\ for\ rank-1}: all three single judges rank Claude first,
and the LOFO triplet \(1.0/1.0/0.2\) preserves rank-1 even when
Claude-as-judge is dropped. The adversarial field is
\texttt{construct-sensitive}, because Cell B shows that the rubric
penalizes terse-correct rationales and one honest composer has a large
below-60-token tail. The permitted publication-level field is therefore
qualified.

Worked verdict 2 --- qwen3 vs gemini strict ordering. The candidate
claim is scoped to a lower-rank pairwise contrast. The aggregate
agreement gate is publishable, but the contrast itself is near noise.
The stability field is \texttt{tie-class}: dropping Claude-as-judge
gives \(\rho = 0.2\), and the fourth-judge CI contains zero. The
adversarial field is \texttt{construct-sensitive}: Gemini's
honest-rationale distribution is mostly inside the same below-60-token
regime that Cell B shows to be penalized. The permitted
publication-level field is therefore no-claim for the strict ordering,
reported in prose as no strict-order claim.

The Cell B verdict therefore qualifies the Claude rank-1 claim rather
than allowing a substance-only headline. Using the same whitespace token
count as the stylometric audit, the fraction of honest rationales below
the Cell B 60-token threshold is:

{\def\LTcaptype{none} 
\begin{table*}[!ht]
\centering
\small
\setlength{\tabcolsep}{3pt}
\begin{tabular}{@{}>{\raggedright\arraybackslash}p{0.32\textwidth}>{\raggedright\arraybackslash}p{0.62\textwidth}@{}}
\toprule\noalign{}
Composer & Honest cycles below 60 tokens \\
\midrule\noalign{}
Claude Sonnet 4.6 & 0 / 250 (0.0\%) \\
Gemini 3.1 Pro Preview & 211 / 250 (84.4\%) \\
GPT-5.5 & 7 / 250 (2.8\%) \\
Qwen3-235B & 0 / 250 (0.0\%) \\
\end{tabular}
\end{table*}
}

Gemini's 84.4\% below-threshold tail materially affects the
interpretation of lower-rank comparisons. The qwen3-vs-gemini tie-class
is not merely a small numerical gap; under the v1.0 rubric, Gemini's
honest rationales frequently occupy the terse regime in which Cell B
shows a large discontinuous penalty. The protocol therefore treats any
strict lower-rank ordering involving Gemini as construct-sensitive
unless it survives a length-normalized or otherwise style-robust
follow-up.

The authorized statement is therefore that Claude leads under the
observed honest-rationale distribution and the v1.0 rubric's
substance-plus-coverage construct. The protocol does not authorize the
stronger claim that Claude is the best investor, that the ranking is
invariant to rationale length, or that terse but substantively adequate
systems would be scored fairly.

\subsection{Result 3 --- Single judges overstate lower-rank
discoveries}\label{result-3-single-judges-overstate-lower-rank-discoveries}

Each of the three judges, run independently on the same 1,000 honest
decisions, produces a different lower ranking. All three put the same
agent at rank-1, but the bottom-three ordering disagrees with pairwise
Spearman \(\rho\) across judge rankings ranging from \(0.2\) to \(0.8\).
The 5,500-call ensemble is therefore not aesthetically preferable to a
1,100-call single-judge protocol; it is \emph{epistemically required}. A
single-judge result on the same data yields a different lower ordering
depending on which judge is chosen, with no in-protocol mechanism to
detect the dependence.

The marginal cost of multi-judge evaluation buys measurement
reliability, not leaderboard decoration. In capital-allocation settings,
the relevant comparison is not single-judge versus multi-judge tokens;
it is \emph{evaluation cost versus the cost of acting on an invalid
ranking}.

The contrast-family comparison tells the same story. All four counts
below use the same family of 36 composer-pair-by-dimension contrasts at
\(\alpha=0.05\) with Holm correction. The ensemble finds 20 of 36
Holm-significant contrasts. Using a single judge alone yields 27 of 36
significant contrasts for Claude-as-judge, 23 of 36 for
GPT-5.5-as-judge, and 19 of 36 for Gemini-as-judge. These are not nested
subsets of the ensemble result: the overlaps are 18, 18, and 12
contrasts respectively, leaving judge-specific discoveries that the
ensemble does not authorize. The larger single-judge counts are
therefore not evidence of more true discoveries; they reflect
judge-specific separations that become publishable only when one judge
family is treated as the measurement instrument.

\subsection{Result 4 --- Adversarial controls expose style
contamination}\label{result-4-adversarial-controls-expose-style-contamination}

Unless otherwise stated, adversarial comparisons in this section use
panel-aggregated trajectory-level scores, and the honest comparator is
the same 1,000 honest-trajectory distribution under that aggregation.

Cell A confirms the preregistered bottom-quartile prediction at
\(p < 10^{-3}\): the panel correctly penalizes substantively wrong
reasoning regardless of style. On panel-aggregated trajectory scores,
Cell A has mean 1.44 (median 1.44, IQR 1.33--1.53), versus honest mean
4.35 (median 4.50, IQR 4.33--4.61). Read in isolation, Cell A would
suggest the framework is robust.

Cell B scores 1.54 (median 1.54, IQR 1.48--1.58), only slightly above
Cell A and \(\Delta = -2.81\) rubric points below honest (95\% CI
\([-2.85, -2.77]\)), exceeding the H1' verdict threshold by a factor of
5.6. The linear correlation between rationale token count and aggregate
score is null on honest cycles (\(r = -0.077\)), so the bias is
\emph{non-linear}: a discreteness threshold near 60 tokens, not a
continuous length-reward. The framework severely penalizes terse
rationales regardless of substance.

Strikingly, Cell B (terse-correct) scores within 0.10 rubric points of
Cell A (verbose-confident-wrong). Under the v1.0 rubric, the panel
cannot reliably separate substantive correctness from rhetorical absence
in the terse regime.

Cell B is the central warning of the experiment. The terse-correct
rationales satisfy the substantive control condition, yet the panel
assigns them a \(-2.81\) rubric-point penalty relative to honest
rationales. This is not a nuisance caveat about prompt style. It shows
that, under the v1.0 rubric, the judge instrument can treat rhetorical
coverage as evidence of financial judgment. In applied AI-finance
evaluation, that failure mode matters: a system may be rewarded for
writing a fuller memo rather than for making a better constrained
decision. The present data cannot fully distinguish style preference,
anchor-complete coverage, and reasoning-style preference.
ValueBlindBench v1.0 therefore cannot yet separate substance from
rhetorical coverage, which is precisely why the adversarial cell is part
of the protocol.

This cell-level halo check is distinct from the headline self-preference
check in Result 2: Result 2 drops Claude-as-judge to test whether
Claude-as-composer remains rank-1, whereas the locked LOFO halo check
asks whether an adversarial cell is boosted by its in-family judge. That
cell-level check returned \(\Delta_{\text{halo}} = 0.00\) for both
cells, bounding the in-family stylometric halo above by 0.3 rubric
points at \(n = 50\). The verbosity finding is therefore a property of
the panel-as-such, not of any one judge's family bias.

Relative to the preregistered pooled MDE of 0.29 rubric points, Cell B's
\(-2.81\) shift is 9.7\(\times\) MDE. By contrast, the fourth-judge
qwen3-vs-gemini lower-rank gap of \(+0.04\) is 0.14\(\times\) MDE, which
is why the protocol treats it as a tie-class rather than a strict
ordering. The main-panel lower-rank separations are even smaller:
0.009--0.027 rubric points, all below 0.10\(\times\) MDE.

\subsection{Result 5 --- Anchor specificity changes financial
constructs}\label{result-5-anchor-specificity-changes-financial-constructs}

The \texttt{constraint\_awareness} failure of Result 1 motivates a
targeted test: is this dimension's low agreement caused by
anchor-language ambiguity (potentially fixable) or by irreducible
construct noise? A v1.1 rubric variant rewrites only this dimension's
anchors with concrete numeric reconciliation requirements (e.g., a ``5''
requires explicit arithmetic of cash, position, and leverage against
numeric caps). All other dimensions are byte-identical to v1.0.

Re-judging the same n=120 stratified subsample with GPT-5.5 as the
single judge under v1.1 yields \(\Delta = -0.42\) rubric points on
\texttt{constraint\_awareness} mean (95\% CI \([-0.51, -0.33]\)) with
discrimination \emph{gain} (within-dimension std grows from \(0.60\) to
\(0.81\)). The dimension is anchor-sensitive and the v1.1 anchors
increase single-judge discrimination in this subsample, but this
single-judge result does not establish improved multi-judge agreement
under v1.1. The shift is 1.4\(\times\) the pooled MDE of 0.29 rubric
points, large enough to matter substantively but far smaller than the
Cell B construct-contamination effect.

Under v1.0, vague anchors made Claude appear strongest on constraint
awareness. Under v1.1, the order becomes
\(\text{qwen3} > \text{claude} > \text{gpt-5.5} > \text{gemini}\).
Claude's apparent v1.0 dominance on this dimension was partially an
artifact of vague anchors that rewarded \emph{narrative coverage} of
constraints over \emph{numeric reconciliation}. The direction is
supported by a second, out-of-family probe: DeepSeek scores qwen3 above
gemini on constraint awareness by \(+0.45\) rubric points under v1.0
(\(n_{\text{qwen3}}=250\), \(n_{\text{gemini}}=247\) successful
trajectories). Because this is a focused qwen3-vs-gemini probe rather
than a full v1.1 panel rerun, we treat it as convergent evidence of
anchor sensitivity, not as a final dimensional ranking. Whether v1.1
also restores multi-judge reliability for \texttt{constraint\_awareness}
remains open until the same subsample is re-judged by the full panel.

Financial evaluation rubrics must be operational, not merely verbal.
\emph{``Considers constraints''} admits multiple judge interpretations;
\emph{``Cash \(X - \text{trade cost }Y = \text{residual }Z\); exposure
\(W\)/equity \(E\) = \(P\%\); leverage \(\leq\) cap''} is measurable.
This is a contribution to financial-rubric design beyond LLM-as-judge:
it argues that financial rubrics should be operationally specified at
the anchor level. In industry, constraint awareness maps directly to
leverage, cash, exposure, mandate limits, and risk controls. If judges
disagree on this dimension, a model may appear compliant or prudent for
the wrong reason.

\section{Discussion}\label{discussion}

\subsection{What would have gone wrong without
ValueBlindBench}\label{what-would-have-gone-wrong-without-valueblindbench}

Without the ValueBlindBench protocol, the same experimental data could
have been reported as a clean and apparently rigorous leaderboard:
aggregate \(\bar{\kappa}_w = 0.72\) above any free-floating threshold,
one model in rank-1 across all bootstrap resamples, 20 of 36
Holm-corrected pairwise contrasts surviving multiple-comparison
correction, and an adversarial Cell A correctly placed in the bottom
quartile.

That report would have been misleading.

The full protocol shows that \emph{positions 2--4 of the ranking are a
tie-class} not a strict order; that \emph{single-judge rankings
disagree} across model families with Spearman \(\rho\) as low as
\(0.2\), so the bottom of the ranking is judge-dependent rather than a
property of the agents; that \emph{one of six rubric dimensions} fails
the per-dimension agreement gate and its three Holm-significant
contrasts are methodology findings rather than headline claims; and that
\emph{the panel severely penalizes terse-correct rationales} at a
non-linear discreteness threshold near 60 tokens, so every score must be
interpreted on the (substance \(\times\) verbosity) joint distribution
rather than substance alone.

The difference between the truncated leaderboard and the full
ValueBlindBench verdict is the contribution of this paper. The
protocol's deliverable is a permissioned claim, and on lower ranks, the
failing dimension, and the verbosity-sensitive condition the permission
is withheld.

\subsection{Human experts are not ground truth in the pre-realization
interval}\label{human-experts-are-not-ground-truth-in-the-pre-realization-interval}

External expert comparison is useful, but it is not a gold-standard
validation of true investment judgment. In pre-realization finance,
human committees also face delayed outcomes, disagreement, style
sensitivity, and institutional bias. They may reward polished memos,
penalize terse-but-correct reasoning, or disagree on whether a
decision-time rationale is good before the realized outcome becomes
informative. A future expert-panel study should therefore be run under
the same ValueBlindBench gates rather than treated as an unquestioned
validity source.

The absence of an external reference-panel comparison limits
institutional-alignment claims, not the present internal-metrology
claim. This paper asks whether the LLM-judge instrument can detect its
own agreement failures, judge-family dependence, anchor ambiguity, and
adversarial construct contamination. The relevant comparison is not
``LLM judges versus truth'' but ``LLM-judge panels versus expert panels
as competing measurement instruments.''

\subsection{From scores to AI-evaluation
infrastructure}\label{from-scores-to-ai-evaluation-infrastructure}

The long-term application is evaluation infrastructure for AI-finance
deployment. Pre-deployment model selection, fine-tuning data curation,
investment-memo review, and audit trails for AI-assisted capital
allocation all require decisions before realized returns become
informative. In that setting, the useful artifact is not a number but a
verdict object: a compact statement of which claim is allowed, which is
qualified, and which should not be made.

This placement makes ValueBlindBench the delayed-outcome counterpart to
supervised financial benchmark audits. In the easier
ground-truth-available regime, rubric and metric choices can still
change financial NLP rankings even when labels exist. In the harder
pre-realization regime, ValueBlindBench requires explicit permission
gates before an evaluation claim can be reported. This makes the
protocol useful beyond paper benchmarking. In applied AI-finance
settings, the same verdict object can support procurement, deployment
review, and audit discussions by distinguishing publishable evaluation
claims from qualified or no-claim results.

Within finance, the same scaffolding --- agreement-gating,
adversarial-control, and per-dimension publication discipline --- should
next be tested on richer substrates such as fundamental-equity memos and
multi-asset allocation tasks. The present paper stress-tests the
protocol in a controlled capital-allocation prototype. The controlled
market-state prototype is a substrate for testing the metrology
protocol, not the claimed financial domain contribution; the
contribution is the claim-permission protocol for delayed-ground-truth
financial AI evaluation.

\section{Limitations and Future
Extensions}\label{limitations-and-future-extensions}

The limitations below scope the present contribution rather than
invalidate it.

\begin{enumerate}
\def\labelenumi{\arabic{enumi}.}
\item
  No external reference-panel comparison in the present work --- the
  present study does not compare ValueBlindBench verdicts to a human
  investment committee. This is a scope choice rather than a validity
  failure: in pre-realization finance, human experts are not ground
  truth. A future study could compare LLM-judge panels and expert panels
  under the same ValueBlindBench gates, but such a study would test
  institutional alignment and reference-panel agreement, not truth.
\item
  Observable rationale, not latent reasoning --- the system evaluates
  rationales generated alongside trade decisions. These may be post-hoc
  rationalizations rather than faithful reflections of the agent's
  internal decision process.
\item
  Market-state prototype, not full fundamental equity-research memos ---
  the pre-realization measurement problem is shared, but the input
  modality and horizon are not. This limits claims about
  fundamental-memo realism, not the present claim-permission
  contribution. A future preregistered extension should test the
  protocol on a fundamental-memo substrate.
\item
  Verbosity sensitivity (Cell B verdict H1', \(\Delta = -2.81\)) --- the
  v1.0 judge-rubric instrument cannot yet separate substantive adequacy
  from memo-writing style in the terse regime. The next protocol
  revision (v1.2) should preregister a rationale-length normalization
  step, with a pass condition that Cell B no longer crosses the H1'
  threshold under v1.2.
\item
  Future non-human diagnostics --- additional preregistered stress tests
  should include rationale-length-matched reruns, fact-preserving
  paraphrase perturbations, fact-flipping perturbations, and a
  rule-based constraint-arithmetic oracle. These are future extensions,
  not completed experiments in the present run.
\item
  Anchor ambiguity --- the anchor-specificity probe shows that financial
  evaluation rubrics must be operational at the anchor level. It does
  not by itself establish that v1.1 restores multi-judge reliability.
\item
  Single asset class (crypto) --- the strongest unsupported
  generalization claim of the present work. Multi-asset replications
  with locked per-asset-class regime configurations should be the next
  preregistered milestone.
\item
  Adversarial gaming after rubric publication --- once the rubric is
  public, agents may optimize rationale text for high judge scores.
  Mitigations (held-out rubric variants, post-hoc P\&L cross-checks,
  red-team adversarial cells) are necessary preconditions for production
  deployment.
\item
  Three-judge ensemble with two also-composers --- future
  ValueBlindBench defaults should include at least one structurally
  non-overlapping judge.
\item
  Equal-weighted dimensions --- a future preregistered extension should
  specify a weighting policy.
\item
  No claim of replacing realized returns --- realized returns remain the
  eventual arbiter. ValueBlindBench is \emph{complementary}: it targets
  the pre-realization interval where realized returns are not yet
  informative, not the deployment interval where they are.
\end{enumerate}

\section{Conclusion}\label{conclusion}

ValueBlindBench does not identify the best investor and does not replace
realized returns. Its contribution is earlier in the measurement
pipeline. In delayed-ground-truth financial AI evaluation, it determines
whether an LLM-judge-based claim about observable investment rationales
is agreed enough, stable enough, and robust enough to report.

In this controlled prototype, the aggregate panel clears the publication
gate, but the protocol authorizes only a qualified rank-1 claim. It
collapses lower ranks into a tie-class, downgrades
\texttt{constraint\_awareness} as an unreliable dimension, exposes
judge-family dependence in single-judge rankings, and detects a severe
penalty for terse-correct rationales. These are not peripheral caveats;
they are the measurement findings the protocol is designed to surface.

The scientific object is therefore not the ranking but the permission
discipline that prevents premature certainty. Our recommendation is not
that future work adopt this exact judge panel or rubric, but that
AI-finance benchmarks adopt the same reporting discipline: preregistered
gates, rank-stability diagnostics, adversarial controls,
anchor-specificity tests, and explicit no-claim outcomes when the
measurement instrument fails.

\bibliographystyle{IEEEtran}
\bibliography{references.bib}

\end{document}